\documentclass[11pt,a4paper]{article}
\usepackage[hyperref]{naaclhlt2019}
\usepackage{times}
\usepackage{latexsym}
\usepackage{multirow}
\usepackage{graphicx}
\usepackage{array}
\newcolumntype{L}{>{\arraybackslash}m{12cm}}

\usepackage{url}

\aclfinalcopy 

\title{BARThez: a Skilled Pretrained French Sequence-to-Sequence Model}

\author{Moussa Kamal Eddine \\
  École Polytechnique \\
  \And
  Antoine J.-P. Tixier \\
  École Polytechnique \\\And
  Michalis Vazirgiannis \\
  École Polytechnique \& AUEB \\}

\date{}

\begin{document}
\maketitle

\begin{abstract}
Inductive transfer learning has taken the entire NLP field by storm, with models such as BERT and BART setting new state of the art on countless NLU tasks.
However, most of the available models and research have been conducted for English.
In this work, we introduce BARThez, the first large-scale pretrained seq2seq model for French.
Being based on BART, BARThez is particularly well-suited for generative tasks.
We evaluate BARThez on five discriminative tasks from the FLUE benchmark and two generative tasks from a novel summarization dataset, OrangeSum, that we created for this research. 
We show BARThez to be very competitive with state-of-the-art BERT-based French language models such as CamemBERT and FlauBERT.
We also continue the pretraining of a multilingual BART on BARThez' corpus, and show our resulting model, mBARThez, to significantly boost BARThez' generative performance.
Code, data and models are publicly available.
\end{abstract}

\section{Introduction}
Inductive transfer learning, that is, solving tasks with models that have been pretrained on very large amounts of data, was a game changer in computer vision \cite{krizhevsky2012imagenet}.
In NLP, while annotated data are scarce, raw text is virtually unlimited and readily available.
It thus emerged that the ability to learn good representations from plain text could greatly improve general natural language understanding.

Trained on gigantic amounts of raw data and with hundreds of GPUs, models based on the Transformer architecture \cite{vaswani2017attention}, such as GPT \cite{radford2018improving} and BERT \cite{devlin2018bert}, have set new state-of-the-art performance in every NLU task.
Moreover, users around the world can easily benefit from these improvements, by finetuning the publicly available pretrained models to their specific applications.
This also saves considerable amounts of time, resources and energy, compared with training models from scratch.

BART \cite{lewis2019bart} combined a BERT-liked bidirectional encoder with a GPT-like forward decoder, and pretrained this seq2seq architecture as a denoising autoencoder with a more general formulation of the masked language modeling objectives of BERT.
Since not only BART's encoder but also its decoder is pretrained, BART excels on tasks involving text generation.

While the aforementioned efforts have made great strides, most of the research and resources were dedicated to the English language, despite a few notable exceptions. In this paper, we partly address this limitation by contributing BARThez\footnote{named after a legendary French goalkeeper, Fabien Barthez: \tiny\url{ https://en.wikipedia.org/wiki/Fabien_Barthez}}, the first pretrained seq2seq model for French.

BARThez, based on BART, was pretrained on a very large monolingual French corpus from past research that we adapted to suit BART's specific perturbation schemes.
Unlike already existing BERT-based French language models such as CamemBERT \cite{martin2019camembert} and FlauBERT \cite{le2019flaubert}, BARThez is particularly well-suited for generative tasks.
We evaluate BARThez on five sentiment analysis, paraphrase identification, and natural language inference tasks from the recent FLUE benchmark, and two generative tasks from a novel French summarization dataset, OrangeSum, that we created for this research. 
We show that BARThez is very competitive with CamemBERT, FlauBERT, and mBART.
We also continue the pretraining of an already pretrained multilingual BART on BARThez's corpus.
Our resulting model, mBARThez, significantly boosts BARThez' performance on generative tasks.

Our contributions are as follows:

\noindent $\bullet$ We publicly release the first large-scale pretrained seq2seq model dedicated to the French language, BARThez, featuring 165M parameters, and trained on 101 GB of text for 60 hours with 128 GPUs.
We evaluate BARThez on five discriminative tasks and two generative tasks, with automated and human evaluation, and show that BARThez is very competitive with the state of the art.

\noindent $\bullet$ To address the lack of generative tasks in the existing FLUE benchmark, we put together a novel dataset for summarization in French, OrangeSum, that we publicly release\footnote{\tiny \url{https://github.com/Tixierae/OrangeSum}} and analyze in this paper.
OrangeSum is more abstractive than traditional summarization datasets, and can be considered the French equivalent of XSum \cite{narayan2018don}.

\noindent $\bullet$ We continue the pretraining of a multilingual BART on BARThez' corpus, and show that
our resulting model, named mBARThez, offers a significant boost over BARThez on generative tasks.

\noindent $\bullet$ We publicly release our code and models\footnote{\tiny \url{https://github.com/moussaKam/BARThez}}.
Our models were also integrated into the highly-popular Hugging Face \texttt{\small Transformers} library\footnote{\tiny \url{https://huggingface.co/transformers/model_doc/barthez.html}}.
As such, they can easily be distributed and deployed for research or production within a standard, industrial-strength framework.
They also have their own APIs and can be interactively tested online.

\section{Related work}
Learning without labels is enabled via self-supervised learning\footnote{a term coined by Yann LeCun.}, a setting in which a system learns to predict part of its input from other parts of its input. In practice, one or more supervised tasks are created from the unlabeled data, and the model learns to solve these tasks with custom objectives.

Some of the earliest and most famous self-supervised representation learning approaches in NLP are word2vec
\cite{mikolov2013distributed}, GloVe \cite{pennington2014glove} and FastText \cite{bojanowski2017enriching}.
While these methods were significant advancements, they produce static representations, which is a major limitation, as words have different meanings depending on the unique contexts in which they are used.

\noindent \textbf{Deep pretrained language models}.
ELMo \cite{peters2018deep} provided the first contextualized embeddings, by extracting and combining the internal states of a pretrained deep bi-LSTM language model.
Except for the word embeddings and the softmax layer, the forwards and backwards RNNs have different parameters.
The authors of ELMo showed that the learned representations could be transferred with great benefits to downstream architectures, to solve a variety of supervised NLU tasks.

Beyond simply combining internal states, \citet{howard2018universal} proposed ULMFiT, a universal transfer learning method for text classification where the language model is pretrained on a large, general dataset, finetuned on a specific dataset, and finally augmented with classification layers trained from scratch on downstream tasks.

With the OpenAI GPT, \citet{radford2018improving} capitalized on the Transformer architecture \cite{vaswani2017attention}, superior and conceptually simpler than recurrent neural networks.
More precisely, they pretrained a left-to-right Transformer decoder as a general language model, and finetuned it on 12 language understanding tasks by applying different transformations to the input.

By combining ideas from all the aforementioned models, and introducing bidirectional pretraining, BERT \cite{devlin2018bert} disrupted the NLP field by setting new state-of-the-art performance on 11 NLU tasks, with very wide margins.
More precisely, BERT uses a bidirectional Transformer encoder with a masked language model objective, making the learned representations capture both the left and the right contexts, instead of just the left context.
The sheer size of BERT, with up to 24 Transformer blocks, plays a role in performance too.

With GPT-2, a version of GPT with over an order of magnitude more parameters than GPT, \citet{radford2019language} showed that as long as they have very large capacities, general language models can reach reasonable performance on many specific NLU tasks out-of-the-box, without any finetuning, i.e., accomplish zero-shot transfer.
This demonstrates the fundamental nature and importance of the language modeling objective for inductive transfer learning.

In RoBERTa, \citet{liu2019roberta} showed that the performance of BERT could be improved by optimizing its hyperparameters and training procedure. The study of why and how BERT works so well has now its own dedicated research field, known as BERTology \cite{rogers2020primer}.

\noindent \textbf{Languages}.
Following the success of BERT for the English language, some BERT models were pretrained and evaluated in other languages.
Some examples include Arabic \cite{antoun2020arabert}, Dutch \cite{de2019bertje,delobelle2020robbert}, French \cite{martin2019camembert,le2019flaubert}, Italian \cite{polignano2019alberto}, Portuguese \cite{souza2019portuguese}, Russian \cite{kuratov2019adaptation}, and Spanish \cite{CaneteCFP2020}.

In addition to the aforelisted monolingual models, multilingual models were also proposed, notably mBERT \cite{devlin2018bert}, XLM \cite{lample2019cross} and XLM-R \cite{conneau2019unsupervised}.

\noindent \textbf{Abstractive summarization}.
Abstractive summarization is an important and challenging task, requiring diverse and complex natural language understanding and generation capabilities.
A good summarization model needs to read, comprehend, and write well.

GPT-2 can be used for summarization, by sampling a certain numbers of tokens from a given start seed.
However, while the generated text is grammatical and fluent, summarization performance is only slightly superior to that of a random extractive baseline.

Being a bidirectional encoder, BERT cannot be used out-of-the-box for language generation, unlike GPT-2.
Furthermore, BERT produces single-sentence representations, whereas for summarization, reasoning over multiple sentence and paragraph representations is necessary.
\citet{liu2019text} proposed a way to overcome these challenges.
At the input level, they introduced special tokens to encode individual sentences, interval segment embeddings, and used more position embeddings than in BERT.
Then, they combined a pretrained BERT encoder with a Transformer-based decoder initialized at random and jointly trained the two models with different optimizers and learning rates.

\noindent \textbf{BART and mBART}.
BART \cite{lewis2019bart} is a denoising auto-encoder that jointly pretrains a bidirectional encoder (like in BERT) and a forward decoder (like in GPT) by learning to reconstruct a corrupted input sequence.
Both the encoder and the decoder are Transformers.
Since not only the encoder but also the decoder is pretrained, BART is particularly effective when applied to text generation tasks.

\citet{liu2020multilingual} pretrained a multilingual BART (mBART) on 25 different languages.
They showed that this multilingual pretraining brings significant performance gains on a variety of machine translation tasks. 
MASS \cite{song2019mass} is another multilingual pretrained sequence to sequence model, that learns to predict a masked span in the input sequence.
The main difference between MASS and BART, is that the former only predicts the masked fragment of the sentence, while the latter learns to reconstruct the entire corrupted sentence.
This difference makes MASS less effective in discriminative tasks, given that only the masked span is fed to the decoder \cite{lewis2019bart}.
ProphetNet \cite{yan2020prophetnet} which also adopts the encoder-decoder structure, introduces a new learning objective called future n-gram prediction. This objective reduces overfitting on local correlations by learning to predict the next n-grams (instead of unigrams) at each time step given the previous context.

\section{BARThez} \label{sec:barthez}
Our model is based on BART \citep{lewis2019bart}, a denoising auto-encoder.
It consists of a bidirectional encoder and a left-to-right auto-regressive decoder.

\subsection{Architecture}
We use the BASE architecture, with 6 encoder and 6 decoder layers.
We did not opt for a LARGE architecture due to resource limitations.
Our BASE architecture uses 768 hidden dimensions and 12 attention heads in both the encoder and the decoder.
In total, our model has roughly 165M parameters.
The architecture has two differences compared with the vanilla seq2seq Transformer \citep{vaswani2017attention}.
The first one is the use of GeLUs activation layers instead of ReLUs, and the second is the presence of a normalization layer on top of the encoder and the decoder, following \citet{liu2020multilingual}.
These additional layers help stabilizing the training when using FP16 precision.

\subsection{Vocabulary}
To generate our vocabulary, we use SentencePiece \citep{kudo2018sentencepiece} that implements byte-pair-encoding (BPE) \citep{sennrich2015neural}.
We do not perform any type of pre-tokenization and we fix the size of the vocabulary to 50K sub-words. The SentencePiece model is trained on a 10GB random sample of the pretraining corpus.
We fix the character coverage to 99.95\%. 

\subsection{Self-supervised learning} \label{subsec:pretraining}
We use the same pretraining as in BART.
That is, BARThez learns to reconstruct a corrupted input.
More precisely, the input text is perturbed with a noise function, and the model has to predict it by minimizing the cross-entropy between the predicted and the original text.
Formally, having a set of documents $\{X_1, X_2, ... , X_n\}$ and a noising function $n$, we aim at finding the parameters $\theta$ that minimize:

$$L_{\theta}=-\sum_i\log P(X_i|n(X_i); \theta)$$

Two different types of noise are applied in $n$.
First, we use the \textit{text infilling} scheme, where a number of text spans are sampled and replaced with one [MASK] special token.
The length of the spans is sampled from a Poisson distribution with $(\lambda=3.5)$ and 30\% of the text is masked.
The second perturbation scheme is \textit{sentence permutation}, where the input document, seen as a list of sentences, is shuffled.

Note that here, we follow \citet{lewis2019bart}, who showed that both text infilling and sentence shuffling were necessary to obtain best results.

\subsection{Pretraining corpus} \label{subsec:pretraining_corpus}
We created a version of FlauBERT's corpus \cite{le2019flaubert} suitable for the two perturbation schemes described in subsection \ref{subsec:pretraining}.
Indeed, in the original FlauBERT corpus, each sentence is seen as an independent instance, while in our case, we need instances to correspond to complete documents.

Other than that, BARThez' corpus is similar to FlauBERT's. 
It primarily consists in the French part of CommonCrawl, NewsCrawl, Wikipedia and other smaller corpora that are listed in Table \ref{table:datasets}.
To clean the corpus from noisy examples, we used the script\footnote{https://github.com/getalp/Flaubert} provided by \citet{le2019flaubert}.
Note that we disabled the Moses tokenizer, as we used SentencePiece which does not require any pre-tokenization.
The total corpus size was 66/101GB before/after SentencePiece tokenization.

\begin{table}[ht]
  \begin{center}
  \small
    \begin{tabular}{|l|c|}
    \hline
      \textbf{Corpus} & \textbf{Size}\\
      \hline
      CommonCrawl & 42.0 \\
      NewsCrawl \cite{li2019findings} & 9.6 \\
      Wikipedia & 4.0 \\
      GIGA \cite{tiedemann2012parallel} & 3.8 \\
      ORTOLANG \cite{11403/est_republicain/v4} & 2.7\\
      MultiUn \cite{eisele2010multiun} & 2.2\\
      EU Bookshop \cite{skadicnvs2014billions}  & 2.1\\
      \hline
    \end{tabular}
  \end{center}
  \caption{BARThez pretraining corpus breakdown (sizes in GB, after cleaning). \label{table:datasets}}
\end{table}

\subsection{Training details}
We pretrained BARThez on 128 NVidia V100 GPUs. We fixed the batch size to 6000 tokens per GPU and the update frequency to 2, which gave a total number of roughly 22k documents per update.
We used the Adam optimizer \cite{kingma2014adam} with $\epsilon=10^{-6}$, $\beta_1=0.9$, and $\beta_2=0.999$, with a learning rate starting from $6.10^{-4}$ and decreasing linearly as a function of the training step.
We used a warm up of 6\% of the total number of training steps.
Pretraining lasted for approximately 60 hours, allowing for 20 passes over the whole corpus.
In the first 12 epochs, we fixed the dropout to 0.1, for epochs 12 to 16 we decreased it to 0.05, and finally we set it to zero for epochs 16 to 20.
All experiments were carried out using the \texttt{Fairseq} library \cite{ott2019fairseq}. 

\begin{table*}[ht]
\small
  \begin{center}{
  \begin{tabular}{| l | c | c c | c c | c c|}
    \hline
    \multirow{2}{*}{\textbf{Dataset}} & \multirow{2}{*}{\textbf{train/val/test}} & \multicolumn{2}{c|}{\textbf{avg. doc length}} & \multicolumn{2}{c|}{\textbf{avg. summary length}} & \multicolumn{2}{c|}{\textbf{vocab size}}\\
    & & \textbf{words} & \textbf{sentences} & \textbf{words} & \textbf{sentences} & \textbf{docs} & \textbf{summaries}\\ \hline 
    CNN &90.3/1.22/1.09&760.50&33.98&45.70&3.58&34&89 \\
    DailyMail &197/12.15/10.40&653.33&29.33&54.65&3.86&564&180 \\
    NY Times &590/32.73/32.73&800.04&35.55&45.54&2.44&1233&293 \\
    \hline
    XSum & 204/11.33/11.33&431.07&19.77&23.26&1.00&399&81 \\
    OrangeSum Title & 30.6/1.5/1.5 & 315.31 & 10.87 & 11.42 & 1.00 & 483 & 43 \\
    OrangeSum Abstract & 21.4/1.5/1.5 & 350 & 12.06 & 32.12 & 1.43 & 420 & 71 \\
    \hline
  \end{tabular}}
  \end{center}
  \caption{Sizes (column 2) are given in thousands of documents. Document and summary lengths are in words. Vocab sizes are in thousands of tokens. \label{table:orangesum_1}}
\end{table*}

\begin{table*}[ht]
\small
  \begin{center}{
  \begin{tabular}{| l | c c c c | c c c | c c c |}
    \hline
    \multirow{2}{*}{\textbf{Dataset}} & \multicolumn{4}{c|}{\textbf{\% of novel n-grams in gold summary}} & \multicolumn{3}{c|}{\textsc{\textbf{lead}}} & \multicolumn{3}{c|}{\textsc{\textbf{ext-oracle}}} \\
    & \textbf{unigrams} & \textbf{bigrams} & \textbf{trigrams} & \textbf{4-grams} & \textbf{R-1} &\textbf{ R-2} & \textbf{R-L} & \textbf{R-1} & \textbf{R-2} & \textbf{R-L} \\ \hline 
    CNN & 16.75 & 54.33 & 72.42 & 80.37 & 29.15 & 11.13 & 25.95 & 50.38 & 28.55 & 46.58 \\
    DailyMail & 17.03 & 53.78 & 72.14 & 80.28 & 40.68 & 18.36 & 37.25 & 55.12 & 30.55 & 51.24 \\
    NY Times & 22.64 & 55.59 & 71.93 & 80.16 & 31.85 & 15.86 & 23.75 & 52.08 & 31.59 & 46.72 \\
    \hline
    XSum & 35.76 & 83.45 & 95.50 & 98.49  & 16.30 & 1.61  & 11.95  & 29.79  & 8.81  & 22.65  \\
    OrangeSum Title & 26.54 & 66.70 & 84.18 & 91.12 & 19.84 & 08.11  & 16.13  & 31.62  & 17.06  & 28.26  \\
    OrangeSum Abstract & 30.03 & 67.15 & 81.94 & 88.3 & 22.21 & 07.00 & 15.48 & 38.36 & 20.87 & 31.08 \\
	\hline
  \end{tabular}}
  \end{center}
\caption{Degree of abstractivity of OrangeSum compared with that of other datasets, as reported in \citet{narayan2018don}.
It can be observed that XSum and OrangeSum are more abstractive than traditional summarization datasets. \label{table:orangesum_2}}
\end{table*}

\begin{table*}[ht]
\centering
\small
\begin{tabular}{|cr|L|}
\hline 
~ & Document &Le 18 octobre dernier, Jacline Mouraud se faisait connaître en publiant sur Facebook une vidéo dans laquelle elle poussait un ``coup de gueule'' contre le gouvernement. Aujourd'hui, la Bretonne a pris ses distances par rapport au mouvement, notamment face à d'autres figures plus radicales comme Éric Drouet. Jacline Mouraud réfléchit désormais à créer son propre parti, ``la seule chose envisageable'', comme elle l'explique au JDD. Nicolas Sarkozy, ``le seul qui a des couilles''. Cette figure des ``gilets jaunes'', accusée de faire le jeu de LREM estime que ``le problème'' d'Emmanuel Macron ``c'est qu'il est jeune''. ``Il devrait y avoir un âge minimum pour être président : 50 ans", souligne Jacline Mouraud. Dans le JDD, elle raconte d'ailleurs avoir voté blanc lors de la dernière présidentielle. En 2007 et 2012, c'est Nicolas Sarkozy, ``le seul qui a des couilles'', que la figure des ``gilets jaunes'' avait soutenu. En attendant de se lancer, pas question pour elle en tous les cas d'être candidate aux européennes sur une liste de La République en marche. \\ 
\hline 
\hline 
\multirow{5}{*}[-2.5em]{\rotatebox[origin=c]{90}{\textsc{Abstract}}} & Gold & L'une des figures du mouvement ne sera toutefois pas candidate aux prochaines élections européennes. \\ 
& mBART & Jacline Mouraud, figure des ``gilets jaunes'', estime que le président \textcolor{orange}{d'}Emmanuel Macron est trop jeune pour être président.\\ 
& mBARThez &Dans un entretien au JDD, la figure des ``gilets jaunes'' Jacline Mouraud révèle qu'elle réfléchit à créer son propre parti. \\ 
& BARThez &Dans les colonnes du JDD, la figure des ``gilets jaunes'' explique qu'elle \textcolor{orange}{envisage} de se présenter aux européennes sur une liste La République en marche.\\ 
& C2C & \textcolor{orange}{Retirée de la vie politique} depuis plusieurs mois, \textcolor{orange}{Bretone} Mouraud envisage de se lancer en politique. Et elle réfléchit à quelque chose de plus, rapporte le JDD.\\ 
\hline 
\hline 
\multirow{5}{*}[-0.1em]{\rotatebox[origin=c]{90}{\textsc{Title}}} & Gold &``Gilets jaunes'' : Jacline Mouraud réfléchit à créer son parti \\ 
& mBART &``Gilets jaunes'' : Jacline Mouraud lance son propre parti\\ 
& mBARThez &``Gilets jaunes'' : Jacline Mouraud prend ses distances\\ 
& BARThez &La figure des ``gilets jaunes'' Jacline Mouraud va créer son propre parti\\ 
& C2C & "Gilets jaunes" : Jacline Mouraud réfléchit à sa propre candidature\\ 
\hline 
\end{tabular} 
\caption{Doc 19233 from OrangeSum's test set, and associated summaries. Incorrect information in \textcolor{orange}{orange}. C2C stands for CamemBERT2CamemBERT. \label{table:example}}
\end{table*}

\section{mBARThez}\label{sec:mbarthez}
mBART \cite{liu2020multilingual} is a multilingual BART.
It follows a LARGE architecture, with 12 layers in both the encoder and the decoder, hidden vectors of size 1024, and 16 attention heads.
It was trained on a multilingual corpus containing 1369 GB of raw text, for over 2.5 weeks on 256 Nvidia V100 GPUs.
The multilingual corpus covers 25 different languages, including 56 GB of French text.
In the original paper, the authors evaluated mBART on machine translation.
However, mBART can also be used to perform monolingual tasks.

We continued the pretraining of the pretrained mBART on BARThez' corpus (see subsection \ref{subsec:pretraining_corpus}) for about 30 hours on 128 Nvidia V100 GPUs, which allowed for 4 passes over BARThez' corpus.
This can be seen as an instance of \textit{language-adaptive pretraining}, which goes a step further than \textit{domain-adaptive pretraining} \cite{gururangan2020don}.
The initial learning rate was set to 0.0001 and linearly decreased towards zero.
We call the resulting model mBARThez.

Note that being multilingual, mBART uses a vocabulary containing tokens with non-latin characters.
We eliminated these tokens from all embedding layers of mBARThez, reducing its number of parameters from 610M to 458M.

\section{OrangeSum}
BART-based models are particularly well-suited to generative tasks, but unfortunately, FLUE \cite{le2019flaubert}, the French equivalent of GLUE, only contains discriminative tasks\footnote{There is no generative task in GLUE or superGLUE \cite{wang2019superglue} either.} \cite{wang2018glue}.

We therefore decided to create one such task.
We opted for single-document abstractive summarization, as it is a generative task that also requires the model to encode its input very well.
In other words, for a model to summarize well, it needs to both read, comprehend, and write well, making abstractive summarization one of the most central and challenging evaluation tasks in NLP.

\noindent \textbf{Motivation}. Our strategy here was to create a French equivalent of the recently introduced XSum dataset \cite{narayan2018don}.
Unlike the historical summarization datasets, CNN, DailyMail, and NY Times, introduced by \citet{hermann2015teaching}, which favor extractive strategies, XSum requires the models to display a high degree of abstractivity to perform well.
XSum was created by scraping articles and their one-sentence summaries from the BBC website, where the one-sentence summaries are not catchy headlines, but rather capture the gist of the articles.

\noindent \textbf{Data collection}.
We adopted an analogous strategy, and scraped the ``Orange Actu'' website\footnote{\url{https://actu.orange.fr/}, `Actu' means News.}.
Orange S.A. is a large French multinational telecommunications corporation, with 266M customers worldwide.
Our scraped pages cover almost a decade from Feb 2011 to Sep 2020.
They belong to five main categories: France, world, politics, automotive, and society\footnote{root URLs are \url{https://actu.orange.fr/} for all categories except \url{https://auto.orange.fr/news/} for automotive.}.
The society category is itself divided into 8 subcategories: health, environment, people, culture, media, high-tech, unsual (``insolite'' in French), and miscellaneous.

Each article featured a single-sentence title as well as a very brief abstract, both professionally written by the author of the article.
We extracted these two fields from each page, thus creating two summarization tasks: OrangeSum Title and OrangeSum Abstract.
Gold summaries are respectively 11.42 and 32.12 words in length on average, for these two tasks (see Table \ref{table:orangesum_1}).
Note that like in XSum, titles in OrangeSum tend not to be catchy headlines but rather convey the essence of the article.
The same can be said about the abstracts.

\noindent \textbf{Post-processing}.
As a post-processing step, we removed all empty articles, and articles whose titles were shorter than 5 words.
For OrangeSum Abstract, we removed the top 10\% articles in terms of proportion of novel unigrams in the abstracts, as we observed that such abstracts tended to be introductions rather than real abstracts.
This corresponded to a threshold of 57\% novel unigrams.

For both OrangeSum Title and OrangeSum Abstract, we set aside 1500 pairs for testing, 1500 for validation, and used all the remaining ones for training.
We make the dataset publicly available\footnote{https://github.com/Tixierae/OrangeSum}.

An example document with its summaries is provided in Table \ref{table:example}.
More examples are available in appendix.

\noindent \textbf{Analysis}.
Table \ref{table:orangesum_1} compares OrangeSum with XSum and the well-known CNN, DailyMail, and NY Times datasets.
We can see that the two OrangeSum datasets are very similar to XSum in terms of statistics, but is one order of magnitude smaller than XSum.
However, the size of OrangeSum still allows for effective finetuning, as we later demonstrate in our experiments.

Table \ref{table:orangesum_2} provides empirical evidence showing that like XSum, OrangeSum is less biased towards extractive systems compared with the traditional datasets used for abstractive summarization.
There are 30\% novel unigrams in the OrangeSum Abstract reference summaries and 26.5\% in OrangeSum Title, compared with 35.7\% in Xsum, 17\% in CNN, 17\% in DailyMail, and 23\% in NY Times.
This indicates that XSum and OrangeSum summaries are more abstractive.
These observations are also confirmed by the fact that the two extractive baselines \textsc{LEAD} and \textsc{EXT-ORACLE} perform much more poorly on XSum and OrangeSum than on the other datasets.

\begin{table*}[ht] 
\centering
\small
\begin{tabular}{|cl|c|c|c|c|c|c|}
\hline
  & & \textbf{Layers} & \textbf{Params} & \textbf{Vocab. size} & \textbf{Pretraining hours} & \textbf{Pretraining GPUs} & \textbf{Corpus size}
\\
  \hline
    \multirow{2}{*}{\rotatebox[origin=c]{90}{\tiny \textsc{BASE}}} 
   & BART-random & 12 & 165 & 50 & 0 & NA & NA \\
&  BARThez (ours) & 12 & 165 & 50 & 60 & 128 & 66 \\
\hline
    \multirow{3}{*}{\rotatebox[origin=c]{90}{\tiny \textsc{LARGE}}} 
 & C2C & 24 & 274 & 32 & 24 & 256 & 138 \\
 & mBART & 24 & 610 & 250 & 432 & 256 & 1369 \\
&  mBARThez (ours) & 24 & 458 & 101 & 30 & 256 + 128 & 1369 + 66 \\
  \hline
\end{tabular}
\caption{Summary of the models used in our experiments. 
Parameters are given in millions, vocab sizes in thousands, and corpus sizes in GB.
C2C stands for CamemBERT2CamemBERT.}
\label{table:summary_models}
\end{table*}

\section{Experiments}
We compare BARThez and mBARThez with the following models, summarized in Table \ref{table:summary_models}.

\noindent $\bullet$ \textbf{mBART}.
The multilingual BART LARGE described in section \ref{sec:mbarthez}.

\noindent $\bullet$ \textbf{CamemBERT2CamemBERT (C2C)}.
To apply CamemBERT to our generative task, we used the BERT2BERT approach  proposed by \citet{rothe2020leveraging}.
More precisely, we fine-tuned a sequence-to-sequence model whose both encoder and decoder parameters were initialized with CamemBERT LARGE weights.
The only weights that were initialized randomly are the encoder-decoder attention weights.

\noindent $\bullet$ \textbf{BART-random}.
As an additional baseline, we train a model with the same architecture and vocabulary as BARThez from scratch on the downstream tasks.

\begin{table*}[t] 
\centering
\small
\begin{tabular}{|cl|cccc|cccc|} 
\hline
   & & \multicolumn{4}{c|}{\textbf{Abstract}} & \multicolumn{4}{c|}{\textbf{Title}} \\
  & & R-1 & R-2 & R-L & BertScore & R-1 & R-2 & R-L & BertScore
\\
  \hline
    \multirow{2}{*}[-2.5em]{\rotatebox[origin=c]{90}{\tiny \textsc{BASE}}} 
 & LEAD & 22.21 & 07.00 & 15.48 & 14.66/68.02 & 19.84 & 08.11 & 16.13 & 15.75/68.43 \\
 & EXT-ORACLE &38.36&20.87&31.08&28.99/73.39&31.62&17.06&28.26&25.15/71.95 \\
  \hline
  \hline
  \multirow{3}{*}[-2.35em]{\rotatebox[origin=c]{90}{\tiny \textsc{LARGE}}} 
 & BART-random &27.67&08.23&18.50&22.53/70.97&28.76&13.15&25.20&29.67/73.65 \\
&  BARThez (ours) &\textbf{31.44}&\textbf{12.77}&\textbf{22.23}&\textbf{27.51}/\textbf{72.84}&\textbf{40.86}&\textbf{23.68}&\textbf{36.03}& \textbf{40.61}/\textbf{77.74}\\
  \hline
 & CamemBERT2CamemBERT &29.23&09.79&19.95&25.53/72.09&34.92&18.04&30.83&36.40/76.17 \\
 & mBART &31.85&13.10&22.35 & 27.80/72.94&40.74&23.70&36.04&40.42/77.67 \\
&  mBARThez (ours) &\textbf{32.67}&\textbf{13.73}&\textbf{23.18}&\textbf{28.80/73.32}&\textbf{41.08}&\textbf{24.11}&\textbf{36.41}&\textbf{41.42/78.05} \\
  \hline
\end{tabular}
\caption{Results on OrangeSum.
The two BertScore scores are with/without rescaling \cite{zhang2019bertscore}.
}
\label{table:eval_summarization}
\end{table*}

\begin{table*}[t] 
\centering
\small
\begin{tabular}{|cl|cccc|cccc|} 
\hline
  & & \multicolumn{4}{c|}{\textbf{OrangeSum Abstract}} & \multicolumn{4}{c|}{\textbf{OrangeSum Title}} \\
 &  & 1-grams & 2-grams & 3-grams & 4-grams & 1-grams & 2-grams & 3-grams & 4-grams
\\
  \hline
 & Gold & 30.03 & 67.15 & 81.94 & 88.30 & 26.54 & 66.70 & 84.18 & 91.12 \\
  \hline
  \hline
{\tiny \textsc{BASE}} & BARThez (ours) & 10.93 & 34.03 & 47.97 & 56.80 &16.69&51.70&72.05&82.49\\
\hline
    \multirow{3}{*}{\rotatebox[origin=c]{90}{\tiny \textsc{LARGE}}} &  C2C & 39.43 & 79.12 & 92.04 & 96.28 & 33.82 & 75.74 & 91.77 & 96.71 \\
    & mBART  & 13.40 & 38.94 & 53.69 & 62.61 &16.89 &52.28&73.12&82.74 \\
 & mBARThez (ours) &\textbf{15.48}&\textbf{43.34}&\textbf{58.53}&\textbf{67.30}&\textbf{17.79}&\textbf{53.41}&\textbf{73.38}&\textbf{82.94} \\
  \hline
\end{tabular}
\caption{Proportion of novel n-grams in the generated summaries. C2C stands for CamemBERT2CamemBERT.
Note that C2C's high scores are misleading as many of the introduced words are irrelevant.}
\label{table:novel_ngrams}
\end{table*}

\begin{table}[ht] 
\centering
\small
\begin{tabular}{|crcc|} 
\hline 
&  & \textbf{Length} & \textbf{Repetitions} (\%) \\ 
\hline 
\multirow{5}{*}[-0.1em]{\rotatebox[origin=c]{90}{\textsc{Abstract}}} & Gold & 32.12 & 11.47   \\
& mBART & 28.20 & 7.47  \\
& mBARThez & 29.45 & 8.60   \\
& BARThez & 29.10 & 14.47  \\
& C2C & 30.68 & 23.00 \\ 
\hline
\hline 
\multirow{5}{*}[-0.1em]{\rotatebox[origin=c]{90}{\textsc{Title}}} & Gold & 11.42 & 0.93  \\
& mBART & 10.79 & 1.73   \\
& mBARThez & 11.03 & 2.27  \\
& BARThez & 11.19 & 2.73  \\
& C2C & 11.23 & 19.53   \\ 
\hline 
\end{tabular} 
\caption{Summary statistics.\label{table:summary_stats}} 
\end{table}

\begin{table}[ht]
  \begin{center}
  \small
    \begin{tabular}{|cl|c|}
    \hline
      & \textbf{System} & \textbf{Score}\\
      \hline
       \multirow{3}{*}[-2.5em]{\rotatebox[origin=c]{90}{\tiny \textsc{LARGE}}} & Gold & 14.29 \\
      \hline
      \hline
      \textsc{\tiny BASE} & BARThez (ours) & \textbf{21.43} \\
      \hline
    & CamemBERT2CamemBERT & -75.00 \\
      & mBART & 11.90 \\
      & mBARThez (ours)  & \textbf{27.38} \\
      \hline
    \end{tabular}
  \end{center}
  \caption{Human evaluation using Best-Worst Scaling.} \label{table:human_eval}
\end{table}

\begin{table*}[t] 
\centering
\small
\begin{tabular}{|cl|c|c|c|c|c|}
\hline
   & & \textbf{CLS-books} & \textbf{CLS-DVD} & \textbf{CLS-music} & \textbf{PAWSX} & \textbf{XNLI}
\\
  \hline
   \multirow{5}{*}{\rotatebox[origin=c]{90}{\small \textsc{BASE}}}
  & mBERT$^\dagger$ \cite{devlin2018bert} & 86.15 & 89.90 & 86.65 & 89.30 & 76.9 \\
  & CamemBERT$_{BASE}$$^\dagger$ \cite{martin2019camembert}  & 92.30 & 93.00 & 94.85 & \textbf{90.14} & \textbf{81.20} \\
  & FlauBERT$_{BASE}$$^\dagger$ \cite{le2019flaubert} & 92.30 & 92.45 & 94.10 & 89.49 & 80.60 \\
& BARThez (ours) & \textbf{94.47}$_{0.17}$ & \textbf{93.17}$_{0.40}$ & \textbf{94.97}$_{0.25}$ & 88.90$_{0.24}$ & 80.73$_{0.40}$ \\
  & BART-random & 76.37$_{0.34}$ & 73.20$_{0.65}$ & 76.00$_{1.28}$ & 55.27$_{0.33}$ & 60.43$_{0.87}$ \\
  \hline
   \multirow{4}{*}{\rotatebox[origin=c]{90}{\small \textsc{LARGE}}}
  & Camembert$_{LARGE}$ \cite{martin2019camembert} & \textbf{95.47}$_{0.33}$ & \textbf{95.37}$_{0.07}$ & \textbf{96.00}$_{0.29}$ & \textbf{91.83}$_{0.54}$ & \textbf{85.33}$_{0.05}$ \\
  & Flaubert$_{LARGE}^\dagger$ \cite{le2019flaubert} & 95.00 & 94.10 & 95.85 & 89.34 & 83.40 \\
  & mBART \cite{liu2020multilingual} & 93.40$_{0.22}$ & 93.10$_{0.20}$ & 93.13$_{0.79}$  & 89.70$_{0.22}$ & 81.07$_{0.38}$  \\
  & mBARThez (ours) & 94.63$_{0.05}$ & 94.03 $_{0.09}$ &  95.30$_{0.16}$ &  90.90$_{0.22}$ & 81.87$_{0.50}$ \\
  \hline
\end{tabular}
\caption{Accuracy on discriminative tasks.
We report the average accuracy over 3 runs, with standard deviation as subscript. $^\dagger$ are taken from \citet{le2019flaubert}.}
\label{table:eval_FLUE}
\end{table*}

\subsection{Summarization}
All pretrained models were finetuned for 30 epochs and we used a learning rate that warmed up to 0.0001 (6\% of the training steps) and then decreased linearly to 0. BART-random was trained for 60 epochs.
We selected the checkpoint associated with the best validation score to generate the test set summaries, using beam-search with a beam size of 4.

We classically report ROUGE-1, ROUGE-2 and ROUGE-L scores \cite{lin2004rouge} in Table \ref{table:eval_summarization}.
However, since ROUGE is limited to capturing n-gram overlap, which is poorly suited to the abstractive summarization setting, we also report BERTScore scores.
BERTScore \cite{zhang2019bertscore} is a recently introduced metric that leverages the contextual representations of the candidate and gold sentences. 

Following \citet{narayan2018don}, we included two extractive baselines in our evaluation, \textsc{LEAD} and \textsc{EXT-ORACLE}.
\textsc{LEAD} creates a summary by extracting the first n sentences from the document.
In our case, we set $n=1$.
The second baseline, EXT-ORACLE, extracts from the document the set of sentences that maximizes a specific score. In our case, we extracted the one sentence maximizing ROUGE-L.

\noindent \textbf{Quantitative results}.
Table \ref{table:eval_summarization} compares the performance of the models finetuned on the summarization task.
While having four times less parameters, BARThez is on par with mBART, both in terms of ROUGE and BERTScore.
mBARThez provides a significant boost over BARThez and mBART and reaches best performance everywhere.
This highlights the importance of adapting a multilingual pretrained model to a specific language before finetuning (\textit{language-adaptive pretraining}).
This also suggests that, when proper adaptation is conducted, it can be advantageous to capitalize on a multilingual model to perform monolingual downstream tasks, probably because there are some translingual features and patterns to be learned.
Finally, all BART-based models outperform CamemBERT2CamemBERT by a significant margin.

\noindent \textbf{Human evaluation}.
To validate our positive quantitative results, we conducted a human evaluation study with 11 French native speakers.
Following \citet{narayan2018don}, we used \textit{Best-Worst Scaling} \cite{louviere2015best}.
In this approach, two summaries from two different systems, along with their input document, are presented to a human annotator who has to decide which one is \textit{better}.
We asked evaluators to base their judgments on three criteria: \textit{accuracy} (does the summary contain accurate facts?), \textit{informativeness} (does the summary capture the important information in the document?) and \textit{fluency} (is the summary written in well-formed French?).

We included the BARThez, mBARThez, mBART and C2C models in our analysis, along with the ground-truth summaries.
We randomly sampled 14 documents from the test set of OrangeSum Abstract, and generated all possible summary pairs for each document, resulting in 140 pairs.
Each pair was randomly assigned to three different annotators, resulting in 420 evaluation tasks in total.
The final score of a model was given as the percentage of time its summary was chosen as \textit{best} minus the percentage of time it was chosen as \textit{worst}.
Scores are reported in Table \ref{table:human_eval}. 
mBARThez reaches first place, like for the quantitative results, but with an even wider margin.
It is also interesting to note that BARThez, which was on par with mBART quantitatively, significantly outperforms it this time around, in terms of human evaluations.
Note that the negative score of CamemBERT2CamemBERT should be analyzed \textit{in comparison} with the other models.
That is, C2C's summaries were judged to be worse more often than not.

Suprisingly, BARThez and mBARThez' summaries were often judged better than the ground truth ones.
We hypothesize that since the GT summaries are short abstracts written by the authors of the articles, they may be well-written but contain information that is missing from the documents, such as dates.
In such situations, the annotators may consider such information as inaccurate (e.g., due to model \textit{hallucinations}) and favor the other model.

\noindent \textbf{Qualitative results}.
As shown in Table \ref{table:novel_ngrams}, mBARThez is more abstractive than BARThez and mBART, as measured by the proportion of novel n-grams in the generated summaries.
E.g., mBARThez introduces on average 15.48\% of novel unigrams in its summaries for the Abstract task, compared with 10.93 and 13.40 for BARThez and mBART, respectively.
It is interesting to note that despite this superior abstractivity, mBARThez still reaches first place everywhere in terms of the ROUGE metric, which measures n-gram overlap.
We hypothesize that BARThez is less abstractive than mBART and mBARThez due to the fact that it is based on a BASE architecture instead of a LARGE one, and has thus four times less parameters.

Finally, it is also to be noted that CamemBERT2CamemBERT (C2C) introduces many new words, which could be considered a good thing at first.
However, it also repeats itself a lot (see Table \ref{table:summary_stats}) and has low ROUGE, BERTSum, and human evaluation scores.
A manual observation revealed that actually, many of the new words introduced by C2C are irrelevant (see appendix for summary examples).

Also, like in \citet{rothe2020leveraging}, we computed the length of the summaries, and the percentage of summaries with at least one non-stopword repetition. We used as stopwords the 500 most frequent words from the system and gold summaries, across all documents.
As can be seen in Table \ref{table:summary_stats}, for both the Abstract and Title tasks, all models generated summaries of sizes very close to that of the Gold summaries.

In terms of repetitions, the less redundant models, closest to the ground truth, are mBART and mBARThez.
This is especially apparent on the Abstract task, where potential for repetition is greater.
On this task, mBART and mBARThez show less than 9\% repetitions, compared with 14.5 and 23 for BARThez and C2C (resp.), and 11.5 in the references.
C2C is also way more redundant than the other models and far from the reference on the Title task, with 19.5\% repetitions.

\subsection{Discriminative tasks}
In addition to generative tasks, BART-like models can perform discriminative tasks \cite{lewis2019bart}.
In the case of sequence classification, the input sequence is fed to both the encoder and the decoder, and the representation of the last token in the sequence is used by adding a classification head on top of it.
When the input consists of several sentences, these sentences are separated with a special token and pasted together.
We evaluate the different models on five discriminative tasks from the FLUE benchmark\footnote{\tiny\url{https://github.com/getalp/Flaubert/tree/master/flue}} \cite{le2019flaubert}, the French equivalent of GLUE \cite{wang2018glue}.

\noindent $\bullet$ \textbf{CLS}. The Cross-lingual Sentiment analysis dataset \cite{prettenhofer2010cross} is made of Amazon reviews to be classified as positive or negative.
It contains 3 product categories: books, DVD and music.
The train and test sets are balanced and contain 2000 examples (each) per product category.
Following  \citet{le2019flaubert}, we used 20\% of the train set as validation set.

\noindent $\bullet$ \textbf{PAWSX}. The Cross-lingual Adversarial Dataset for Paraphrase Identification \cite{yang2019paws} contains pairs of sentences, and the task is to predict whether they are semantically equivalent.
There are 49401 examples for training, 1992 for development, and 1985 for testing.

\noindent $\bullet$ \textbf{XNLI}. The Cross-lingual NLI corpus \cite{conneau2018xnli} contains pairs of sentences, and the task is to predict whether the first one (premise) entails the second one (hypothesis), contradicts it, or neither entails nor contradicts it (neutral relationship). 392702 pairs are used for training, 2490 for development, and 5010 for testing.

\noindent\textbf{Training details}. In all experiments, we finetune the model for 10 epochs with a learning rate chosen from  \{$10^{-4}$, $5.10^{-5}$, $10^{-5}$\} based on the best validation score.
We repeat each experiment 3 times with different seeds and report the mean and standard deviation.

\noindent\textbf{Results}. Table \ref{table:eval_FLUE} reports the test set accuracies.
For comparison purposes, we also copy that of other relevant BERT-based models as reported in \citet{le2019flaubert}.
These models are mBERT \cite{devlin2018bert}, CamemBERT \cite{martin2019camembert} and FlauBERT \cite{le2019flaubert}.

Among the models having a BASE architecture, BARThez is best in the three sentiment analysis tasks, while being very close to CamemBERT and FlauBERT in the paraphrasing and inference tasks.

Among the LARGE models, mBARThez outperforms mBART in all tasks, showing again the importance of language-adaptive pretraining.
On the other hand, CamemBERT and FlauBERT outperform mBARThez in most of the tasks, which could be attributed to the fact that CamemBERT and FlauBERT were trained for approximately 10 times more GPU hours on a monolingual French corpus.
Nevertheless, given that huge difference in monolingual training time, it is remarkable that mBARThez is so close, and sometimes outperforms, FlauBERT, with e.g., a comfortable 1.56 margin on PAWSX.

We can conclude that the ability of BARThez and mBARThez to perform well on generative tasks does not appear to come at the expense of a decrease in performance on discriminative tasks, which is in line with the results presented in the BART paper \cite{lewis2019bart}.

\section{Conclusion}
We released BARThez and mBARThez, the first large-scale pretrained seq2seq models for the French language, as well as a novel summarization dataset for French, inspired by the XSum dataset.
By evaluating our models on the summarization dataset we showed that:
(1) BARThez is on par with mBART while having four times less parameters, and that (2) mBARThez provides a significant boost over mBART by simply adding a relatively affordable language-adaptive phase to the pretraining.
In addition, we evaluated BARThez and mBARThez on 5 sentiment analysis, paraphrasing, and natural language inference tasks against cutting edge BERT-based French language models (FlauBERT and CamemBERT), and obtained very competitive results.
An interesting area for future work is to further explore the language-adaptive pretraining approach.

\section*{Acknowledgments}
We are thankful to the National Center for Scientific Research (CNRS) for giving us access to their Jean Zay supercomputer, under allocation 2020-AD011011499.

\bibliography{naaclhlt2019}
\bibliographystyle{acl_natbib}

\newpage
\onecolumn

\noindent \textbf{\LARGE Appendix}\\
In what follows, we provide, for 10 documents randomly selected from OrangeSum's test set, the reference and model summaries for each task (Abstract and Title).

\begin{table}[h]
 \centering 
 \begin{tabular}{|cr|L|} 
 \hline 
  & Document & \baselineskip=10pt {\small "Nous pouvons confirmer à ce stade que cinq personnes ont péri. Au moins 70 personnes sont dans un état inconscient dans un hôpital non loin, et dans l'ensemble, entre 200 et 500 personnes reçoivent des soins", a déclaré Swaroop Rani, une responsable de la police de Visakhapatnam, dans l'État de l'Andhra Pradesh.Opérée par LG Polymers, l'usine est située en bordure de la ville industrielle et portuaire de Visakhapatnam. L'agglomération compte une population d'environ 5 millions de personnes.Le gaz "avait été laissé là à cause du confinement. Cela a mené à une réaction chimique et de la chaleur est apparue dans les réservoirs, et le gaz a fui à cause de cela", a expliqué Mme Rani. L'Inde est actuellement confinée depuis fin mars pour lutter contre la pandémie de coronavirus."Nous avons reçu un appel d'urgence de villageois autour de 03H30 du matin aujourd'hui (mercredi 22H00 GMT, ndlr). Ils disaient qu'il y avait du gaz dans l'air", a-t-elle poursuivi."Nous sommes arrivés immédiatement sur place. On pouvait sentir le gaz dans l'air et il ne n'était possible à aucun d'entre nous de rester là plus de quelques minutes", a-t-elle décrit.L'Inde a été le théâtre en décembre 1984 d'un des pires accidents industriels de l'Histoire, lorsque 40 tonnes de gaz s'étaient échappées d'une usine de pesticides de la ville de Bhopal (centre).Quelque 3.500 personnes avaient péri en quelques jours, principalement dans des bidonvilles situés autour de cette usine d'Union Carbide, et des milliers d'autres dans les années qui ont suivi.
  } \\
 \hline 
 \hline 
 \multirow{5}{*}[-3.5em]{\rotatebox[origin=c]{90}{\textsc{Abstract}}} & Gold & Au moins cinq personnes ont péri et plus d'un millier ont été hospitalisées à la suite d'une fuite de gaz dans une usine chimique du sud-est de l'Inde, ont annoncé jeudi la police et une source officielle. \\ 
 & mBART & Cinq personnes sont mortes mercredi dans l'accident d'une usine de pesticides qui s'est produite en Inde, à la suite du confinement lié à l'épidémie de Covid-19, ont indiqué des responsables.\\ 
 & mBARThez &Une explosion a fait cinq morts mercredi dans une usine de pesticides du centre de l'Inde, après que le gaz a fui dans les réservoirs après une réaction chimique, ont indiqué les autorités. \\ 
 & BARThez &Une dizaine de personnes ont péri et des centaines d'autres ont été blessées mercredi dans une usine de pesticides près de Visakhapatnam, dans le sud de l'Inde, a annoncé la police.\\ 
 & C2C & Au moins vingt personnes sont mortes, dont cinq sont mortes et cinq sont portées disparues, selon un bilan officiel lundi après-midi en Inde, faisant craindre une fuite de gaz meurtrière dans le pays, selon une source gouvernementale à l'AFP.\\ 
 \hline 
 \hline 
 \multirow{5}{*}[-2em]{\rotatebox[origin=c]{90}{\textsc{Title}}} & Gold &Fuite de gaz dans une usine en Inde: 5 morts, au moins 1.000 personnes hospitalisées \\ 
 & mBART &Inde: cinq morts dans un accident de la usine de pesticides\\ 
 & mBARThez &Inde: au moins cinq morts dans le crash d'une usine de pesticides\\ 
 & BARThez &Inde: cinq morts dans un glissement de terrain à Visakhapatnam\\ 
 & C2C & Inde: cinq morts dans un gaz mortel dans un usine de recyclage\\ 
 \hline 
 \end{tabular} 
 \caption{C2C stands for CamemBERT2CamemBERT. OrangeSum document 12158.} 
 
 \end{table}

\begin{table*} 
 \centering 
 \begin{tabular}{|cr|L|} 
 \hline 
 & Document &\baselineskip=8pt {\small De nombreux scientifiques occidentaux ont fait part de leurs doutes quant à la rapidité avec laquelle ce vaccin aurait été mis au point.Le ministre américain de la Santé Alex Azar s'est fait l'écho mercredi de leurs points de vue, à l'issue d'une visite de trois jours à Taïwan."Il est important que nous fournissions des vaccins sans danger et efficaces et que les données soient transparentes... Ce n'est pas une course pour être le premier", a-t-il déclaré à la presse lors d'une conférence téléphonique."Je dois souligner que deux des six vaccins américains dans lesquels nous avons investi sont entrés dans la phase des essais cliniques il y a trois semaines, alors que le vaccin russe ne fait que commencer", a-t-il ajouté."Les données des premiers essais en Russie n'ont pas été divulguées, ce n'est pas transparent", a estimé le ministre américain.Mardi, le président russe Vladimir Poutine a annoncé le développement par son pays du "premier" vaccin sans danger contre le Covid-19, affirmant que l'une de ses filles se l'est fait inoculer.Ce vaccin a été baptisé "Spoutnik V" (V comme vaccin, ndlr), en référence à la victoire politico-scientifique russe qu'était la mise en orbite en 1957 du satellite du même nom en pleine Guerre froide.Peu après la déclaration du Kremlin, l'Organisation mondiale de la santé (OMS) a réagi en appelant à la prudence, rappelant que la "pré-qualification" et l'homologation d'un vaccin passaient par des procédures "rigoureuses".De nombreux pays se sont lancés dans une véritable* course contre la montre pour trouver un vaccin efficace contre le coronavirus qui a tué plus de 740.000 personnes à travers la planète depuis son apparition l'an dernier en Chine.Les Etats-Unis sont le pays le touché avec 5,1 millions de cas de Covid-19 qui a fait plus de 164.000 morts. M. Azar s'est dit confiant sur la capacité des Américains à mettre au point un vaccin."Nous pensons qu'il est très crédible que nous ayons des dizaines de millions de doses de vaccin de référence, sûres et efficaces d'ici la fin de cette année, et plusieurs centaines de millions de doses au début de l'année prochaine", a-t-il affirmé.Le président américain a lancé l'opération "Warp Speed" qui vise explicitement à obtenir de quoi vacciner tous les Américains dès janvier 2021.} \\ 
 \hline 
 \hline 
 \multirow{5}{*}[-3.5em]{\rotatebox[origin=c]{90}{\textsc{Abstract}}} & Gold & Le ministre américain de la Santé a fait part mercredi de son scepticisme après l'annonce du développement par la Russie du "premier" vaccin contre le coronavirus assurant une "immunité durable". \\ 
 & mBART & Le ministre américain de la Santé s'est exprimé mercredi sur la possibilité d'un vaccin efficace contre le coronavirus.\\ 
 & mBARThez &Le ministre de la Santé américain Alex Azar a déclaré mercredi que les Etats-Unis et la Russie ont fait des essais cliniques d'un vaccin contre le coronavirus "sans danger", alors que le président russe Vladimir Poutine a déclaré mardi avoir inoculé une de ses filles. \\ 
 & BARThez &Les Etats-Unis sont le pays le plus touché par la pandémie de coronavirus après la Russie qui a pourtant annoncé avoir mis au point le "premier vaccin", a estimé le ministre américain de la Santé.\\ 
 & C2C & Le ministre américain de la Santé a souligné que la Russie avait développé des capacités capables de détecter et de tester si le pays n'était pas vaccin contre le nouveau coronavirus, mais a jugé prudent "dans l'attente de la publication d'une réponse scientifique",\\ 
 \hline 
 \hline 
 \multirow{5}{*}[-2em]{\rotatebox[origin=c]{90}{\textsc{Title}}} & Gold &Coronavirus: le ministre américain de la Santé sceptique au sujet du vaccin russe \\ 
 & mBART &Vaccin " sans danger": les Américains investis en Russie, selon Alex Azar\\ 
 & mBARThez &Vaccin russe: les Américains appelés à la prudence\\ 
 & BARThez &Un vaccin russe contre le Covid-19 en vue aux Etats-Unis, selon le ministre américain de la Santé\\ 
 & C2C & Coronavirus: les Etats-Unis pas en "cours de combattant" face à un vaccin expérimental\\ 
 \hline 
 \end{tabular} 
 \caption{C2C stands for CamemBERT2CamemBERT. OrangeSum document 33555.} 
 
 \end{table*}

\begin{table*} 
 \centering 
 \begin{tabular}{|cr|L|} 
 \hline 
 & Document &\baselineskip=7pt {\scriptsize Une première depuis la Seconde guerre mondiale, la consommation d'alcool ne baisse plus en France. L'Académie nationale de médecine a appelé lundi 29 avril les pouvoirs publics à "prendre des mesures plus fortes" pour lutter contre les problèmes de santé publique causés par la consommation d'alcool. "Pour la première fois depuis la Seconde guerre mondiale, la consommation d'alcool ne baisse plus en France. C'est une défaite majeure pour la santé publique, car l'alcool en est un déterminant fondamental", estime l'Académie dans un communiqué diffusé lundi 29 avril. L'organisme déplore en particulier "l'affaiblissement continu de la loi Evin sous la pression du lobby alcoolier, jusqu'à autoriser la publicité sur l'internet, support médiatique particulièrement affectionné des jeunes". L'alcool serait la première cause évitable* de mortalité des 15-30 ans, selon l'Académie de Médecine. Elle invite donc le gouvernement à revenir aux "principes initiaux" de la loi. Pour un pictogramme plus visible pour les femmes enceintes À l'instar d'autres institutions et associations, l'Académie recommande d'interdire la publicité pour l'alcool et de faire figurer sur les boissons alcoolisées la mention "l'alcool est dangereux pour la santé" (et non le seul excès). L'Académie de médecine veut également voir taxées les boissons au gramme d'alcool et demande la mise en place d'un prix minimum de vente par gramme d'alcool, comme c'est le cas en Ecosse depuis un an. Elle réclame également un pictogramme plus grand et plus lisible sur les bouteilles pour "dissuader de toute consommation la femme enceinte ou qui désire l'être".L'académie de médecine pointe clairement la responsabilité du lobby alcoolier. "Malgré l'enjeu de prévenir la première cause de retard mental évitable* du nouveau-né et de l'enfant, les discussions pour l'agrandir et le contraster s'enlisent depuis des années face à l'opposition farouche du lobby alcoolier". L'alcool serait la première cause de retard mental de l'enfant et de démence précoce souligne l'organisme. Un quart des Français boit trop Dans des chiffres publiés mi-février, Santé publique France avait indiqué que la consommation des Français n'avait quasiment pas reculé depuis 10 ans, passant de 27 g à 26 g d'alcool pur par jour entre 2009 et 2015. "C'est en février 2019 que Santé Publique France annonce que la consommation française d'alcool est la même en 2017 qu'en 2013", note l'académie dans son communiqué. Près d'un quart des Français, soit environ 10,5 millions d'adultes, boivent trop d'alcool, avait également estimé fin mars Santé publique France. L'agence sanitaire a diffusé de nouveaux repères de consommation, résumés par le message "pour votre santé, c'est maximum deux verres par jour, et pas tous les jours".L'alcool constitue la deuxième cause de mortalité évitable* après le tabac, avec 41.000 décès qui lui sont attribuables chaque année en France, 30.000 hommes et 11.000 femmes. L'alcool "est impliqué dans 40\% des violences faites aux femmes et aux enfants et un tiers des décès par accidents de la route", ajoute l'Académie dans son communiqué.} \\ 
 \hline 
 \hline 
 \multirow{5}{*}[-3.5em]{\rotatebox[origin=c]{90}{\textsc{Abstract}}} & Gold & Elle demande des "mesures plus fortes" pour lutter contre les problèmes de santé causés en France par une consommation d'alcool qui ne diminue plus. \\ 
 & mBART & En février 2019, Santé publique France avait indiqué que la consommation des Français n'avait quasiment pas reculé depuis 10 ans.\\ 
 & mBARThez &Près d'un quart des Français boivent trop d'alcool. \\ 
 & BARThez &L'Académie de médecine réclame notamment "l'affaiblissement continu de la loi Evin sous la pression du lobby alcoolier", jusqu'à autoriser la publicité sur l'internet.\\ 
 & C2C & À l'inverse de ce qui se fait en France, la mesure doit inciter à la consommation d'alcool dès l'âge de 18 ans.\\ 
 \hline 
 \hline 
 \multirow{5}{*}[-2em]{\rotatebox[origin=c]{90}{\textsc{Title}}} & Gold &Stagnation de la consommation d'alcool en France : "une défaite majeure pour la santé publique" \\ 
 & mBART &Santé : l'Académie de médecine demande des mesures plus fortes\\ 
 & mBARThez &alcool : l'Académie de médecine appelle le gouvernement à des mesures plus fortes\\ 
 & BARThez &Alcool : une " défaite majeure" pour la santé publique, selon l'Académie de médecine\\ 
 & C2C & La consommation d'alcool en forte hausse : l'Académie de médecine appelle à plus de fermeté\\ 
 \hline 
 \end{tabular} 
 \caption{C2C stands for CamemBERT2CamemBERT. OrangeSum document 25148.} 
 
 \end{table*}

\begin{table*} 
 \centering 
 \begin{tabular}{|cr|L|} 
 \hline 
 & Document &\baselineskip=7pt {\scriptsize De petites dimensions (20 cm de largeur et 30 de hauteur), ces ouvertures à hauteur d'homme percées à côté du porche des somptueux palais appartenant aux grandes familles florentines servaient à écouler le vin directement du producteur au consommateur.Au fil des siècles, ce détail architectural et sa fonction sont tombés dans les oubliettes de l'Histoire jusqu'à ce que Massimo Casprini, un érudit florentin, parte à leur redécouverte et y consacre un livre, "I finestrini del vino" ("Les fenêtres à vin"), publié en 2005.Ces fenêtres "ont été créées à partir de 1532 après la chute de la République, quand les Médicis sont revenus au pouvoir et ont voulu favoriser l'agriculture, incitant les grands propriétaires florentins à investir dans les oliveraies et les vignes (...) tout en leur donnant des avantages fiscaux pour revendre directement leur production en ville", explique à l'AFP M. Casprini lors d'une promenade à travers les rues de Florence dans la touffeur estivale.Unique restriction: "Ils pouvaient y vendre seulement le vin de leur propre production et sous un format particulier d'environ 1,4 litre"."L'autre fonction de ces petites fenêtres était sociale, en permettant aux gens du peuple d'acquérir du vin à prix plus raisonnable que chez les commerçants, sans intermédiaire", ajoute-t-il, précisant dans un sourire qu'"à l'époque la consommation de vin était énorme".- Episodes de peste -A l'heure du coronavirus et de la distanciation sociale, Massimo Casprini rappelle que "grâce à ce système on évitait les contacts", alors qu'"épidémies et épisodes de peste étaient très fréquents au XVIe siècle"."En effet, la fenêtre à vin était fermée par un panneau de bois, le client se présentait et frappait avec le heurtoir, à l'intérieur il y avait un caviste qui prenait la bouteille vide et la remplissait. Il n'y avait donc pas de contact direct!" s'extasie le fringuant septuagénaire, également amateur de motos anciennes et auteur de quelque 70 ouvrages centrés sur la capitale toscane.Jusqu'ici, 267 de ces fenêtres à vin ont été répertoriées en Toscane, dont 149 dans le centre de Florence. "Il y en avait beaucoup plus!" estime M. Casprini, "presque tous les propriétaires terriens avaient une fenêtre à vin, mais nombre d'entre elles ont disparu, notamment lors des bombardements de la Seconde Guerre mondiale".Certaines ont aussi été murées, mais grâce à l'œil de lynx de notre expert on réussit encore à reconnaître les contours de leur encadrement en pietra serena (grès gris) ou pierre des carrières de Fiesole, près de Florence.Dans le fil du livre du professeur Casprini a été fondée une association, baptisée "Le buchette del vino", qui recense et appose une plaque sur chaque fenêtre. Son site internet (https://buchettedelvino.org/) propose même une carte interactive permettant de partir à leur découverte, ainsi qu'une galerie de photos et une présentation historique de ces petits trésors architecturaux.On y cite par exemple un guide en français de Florence datant de 1892 qui mentionne la fenêtre d'un palais: "cette cave assez renommée pour ses vins millésimés ne livre aux consommateurs que ceux provenant des propriétés de la marquise Leonia degli Albizi Frescobaldi".Tombées en désuétude, les "finestrini del vino" font aujourd'hui l'objet d'un regain d'intérêt et d'une forme de recyclage: présentoir de magasin, passe-plat dans un café, ou encore petit autel dédié à la Vierge.Même si elles sont protégées par la loi, M. Casprini déplore que "trois fenêtres ont déjà disparu" depuis son premier recensement en 2005.} \\ 
 \hline 
 \hline 
 \multirow{5}{*}[-3.5em]{\rotatebox[origin=c]{90}{\textsc{Abstract}}} & Gold & Florence, joyau de la Renaissance, peut s'enorgueillir d'un patrimoine mondialement célèbre, mais dont certains détails restent encore aujourd'hui méconnus: c'est le cas des discrètes "fenêtres à vin" ornant la façade de certains palais, qui permettaient la vente de vin "sans contact", un concept redevenu d'actualité en ces temps de coronavirus. \\ 
 & mBART & "A l'heure du coronavirus et de la distanciation sociale, il n'y avait pas de contact direct !" A Florence, des fenêtres à vin, remplacées par des pierres, auraient été oubliées dans lesoubliettes de l'Histoire.\\ 
 & mBARThez &Massimo Casprini, spécialiste des fenêtres à vins, est revenu mercredi sur la révolution de Florence (Italie) où il a redécouvert l'existence de ces ouvertures en plein air et à ciel ouvert qui permettaient, autrefois, à des propriétaires de vins de revendre leur production à la ville. \\ 
 & BARThez &De 1532 à nos jours, les fenêtres à vin des palais anciens de Florence sont les plus souvent murées, un détail qui a sans doute survécu à l'épidémie de nouveau coronavirus.\\ 
 & C2C & Au lieu de la pandémie de coronavirus, un jardin italien a retrouvé des crus du monde entier: ils étaient des caves à vin français, à quelques dizaines de mètres du sol, pour ne pas être contaminés par le Covid-19.\\ 
 \hline 
 \hline 
 \multirow{5}{*}[-2em]{\rotatebox[origin=c]{90}{\textsc{Title}}} & Gold &Virus: comment la Florence des Médicis inventa la vente de vin "sans contact" \\ 
 & mBART &Les fenêtres à vin sont tombées dans les oubliettes de l'Histoire\\ 
 & mBARThez &Florence: les fenêtres à vin cachées dans un livre\\ 
 & BARThez &Florence: des fenêtres à vin traditionnelles à l'heure des peste\\ 
 & C2C & Les fenêtres de la Florence en "huile de vin" : actualité automobile, infos, scoop\\ 
 \hline 
 \end{tabular} 
 \caption{C2C stands for CamemBERT2CamemBERT. OrangeSum document 34657.} 
 
 \end{table*}

\begin{table*} 
 \centering 
 \begin{tabular}{|cr|L|} 
 \hline 
 & Document &\baselineskip=7pt {\scriptsize L'ancien chef de l'État était entendu depuis mardi matin, avec une interruption dans la nuit, dans les locaux de l'office anticorruption (OCLCIFF) situés à Nanterre (Hauts-de-Seine). L'ancien président de l'UMP a regagné son domicile parisien du XVIe arrondissement après la fin de sa garde à vue. Également entendu, mais sous le statut de "suspect libre", Brice Hortefeux, un proche de l'ex-président qui occupa plusieurs postes ministériels pendant le quinquennat Sarkozy (2007-2012), a de son côté quitté les locaux de l'office anticorruption mardi soir, assurant sur Twitter avoir apporté des précisions pour "permettre de clore une succession d'erreurs et de mensonges".Depuis la publication, en mai 2012, par le site d'informations Mediapart d'un document libyen - attribué à l'ex-chef des renseignements Moussa Koussa - accréditant un financement d'environ 50 millions d'euros, les investigations des juges ont considérablement avancé. Plusieurs protagonistes du dossier, dont plusieurs ex-responsables libyens, ont accrédité la thèse de versements illicites. Ziad Takieddine persiste et signeLe sulfureux homme d'affaires Ziad Takieddine a lui-même assuré avoir remis entre fin 2006 et début 2007 trois valises contenant 5 millions d'euros en provenance du régime de Kadhafi à Nicolas Sarkozy, alors ministre de l'Intérieur, et à son directeur de cabinet Claude Guéant. Sur BFMTV, il a réédité ses accusations mais répété que cet argent "n'était pas lié à la campagne présidentielle" de 2007."Cet argent faisait partie des accords entre les deux pays sur le contrôle des frontières maritimes, avec échanges d'informations", a précisé l'homme d'affaires, mis en examen autour de ce dossier pour complicité de corruption et complicité de diffamation. "Il y avait un devoir de former en France des équipes libyennes avant la livraison du matériel. Dans ce cadre-là, il y avait des formations à destination de quelques centaines de Libyens. Ils ont établi en France que ça allait coûter dans les cinq millions d'euros", a-t-il ajouté.Nicolas Sarkozy "est un vrai menteur et vous allez voir, il va passer son temps avec les juges d'instruction à dire 'non, non, non c'est pas vrai'. Tout ça pour gagner du temps, c'est sa méthode habituelle", a également lancé Ziad Takieddine sur BFMTV mercredi, assurant "dire la vérité". L'ancien chef de l'État a toujours rejeté ces mises en cause. D'autres dignitaires libyens ont démenti tout financement de la Libye de Mouammar Kadhafi, que Nicolas Sarkozy avait reçu en grande pompe à l'Élysée en 2007.De nouveaux éléments compromettantsOuverte notamment pour "détournements de fonds publics" et "corruption active et passive", l'enquête a été élargie en janvier à des soupçons de "financement illégal de campagne électorale", suite à un rapport de l'office anticorruption qui pointe une circulation importante d'argent liquide dans l'entourage de Nicolas Sarkozy durant la campagne 2007. Selon Le Monde, plusieurs anciens dignitaires du régime Kadhafi auraient livré de récents témoignages confirmant les soupçons de financement illicite. Les investigations ont aussi mis en lumière un virement de 500.000 euros perçu par Claude Guéant en mars 2008, en provenance d'une société d'un avocat malaisien. L'ex-secrétaire général de l'Élysée a toujours affirmé qu'il s'agissait du fruit de la vente de deux table*aux, sans convaincre les juges qui l'ont mis en examen notamment pour "blanchiment de fraude fiscale en bande organisée".Les juges s'interrogent également sur la vente suspecte en 2009 d'une villa située à Mougins, sur la Côte d'Azur, à un fonds libyen. Ils soupçonnent l'homme d'affaires Alexandre Djouhri d'avoir été le véritable* propriétaire de ce bien et de l'avoir cédé pour 10 millions d'euros, soit plus du double du prix du marché, ce qui aurait pu permettre de dissimuler d'éventuels versements occultes.} \\ 
 \hline 
 \hline 
 \multirow{5}{*}[-3.5em]{\rotatebox[origin=c]{90}{\textsc{Abstract}}} & Gold & Après une vingtaine d'heures, la garde à vue de Nicolas Sarkozy s'est achevée mercredi soir. L'ancien président a été mis en examen pour "corruption passive", "financement illégal de campagne électorale" et "recel de fonds publics libyens" et placé sous contrôle judiciaire dans le cadre de l'enquête sur des soupçons de financement de sa campagne présidentielle de 2007 par la Libye de Mouammar Kadhafi. \\ 
 & mBART & Ziad Takieddine, mis en examen autour de l'affaire des soupçons de financement libyen de la campagne présidentielle de 2007 de Nicolas Sarkozy, a de nouveau quitté les locaux de son établissement, à Nanterre, dans la nuit de mardi à mercredi.\\ 
 & mBARThez &Cinq jours après la révélation d'un document par Mediapart, l'ancien président de l'UMP et principal suspect dans l'affaire des soupçons de financement libyen de sa campagne présidentielle de 2007 a quitté mardi soir les locaux où il était auditionné. Brice Hortefeux et Claude Guéant ont apporté des précisions. \\ 
 & BARThez &L'ancien président de la République Nicolas Sarkozy a quitté mardi matin les locaux de l'office anticorruption où il était entendu. Les soupçons de financement libyen de sa campagne présidentielle de 2007.\\ 
 & C2C & Nicolas Sarkozy est mis en examen dans le cadre de l'enquête sur les soupçons de financement libyen de sa campagne présidentielle de 2007. Selon plusieurs médias, l'ancien chargé de mission a dit mercredi n'être "pas au courant" de ce que l'ex-\\ 
 \hline 
 \hline 
 \multirow{5}{*}[-2em]{\rotatebox[origin=c]{90}{\textsc{Title}}} & Gold &Soupçons de financement libyen : Nicolas Sarkozy mis en examen \\ 
 & mBART &Affaire libyenne : Nicolas Sarkozy en garde à vue\\ 
 & mBARThez &Affaire libyenne : Nicolas Sarkozy entendu par les juges\\ 
 & BARThez &Nicolas Sarkozy en garde à vue, la piste d'un financement libyen s'éloigne\\ 
 & C2C & Nicolas Sarkozy est "un vrai traître" selon l'entourage de Nicolas Sarkozy\\ 
 \hline 
 \end{tabular} 
 \caption{C2C stands for CamemBERT2CamemBERT. OrangeSum document 22208.} 
 
 \end{table*}

\begin{table*} 
 \centering 
 \begin{tabular}{|cr|L|} 
 \hline 
 & Document &Jean-Paul Dufrègne a passé un sale quart d'heure sur les réseaux sociaux mercredi soir. Cet élu communiste de l'Allier a été filmé par les caméras de TF1, dans un reportage diffusé le 4 avril au journal de 20 heures. Mais téléspectateurs et internautes n'ont nullement prêté attention aux arguments du député sur les inquiétudes persistantes des territoires ruraux et la réforme institutionnelle sur laquelle planche le gouvernement. Non, ils étaient bien trop captivés par son compteur de vitesse, filmé le temps de quelques plans par les caméras de la première chaîne, comme le relève LCI.Car, sur une route départementale limitée à 90 km/heure, Jean-Paul Dufrègne avait le pied au plancher. Son compteur affichait 124 km/heure, plus de 30 km/heure au-dessus de la limite autorisée. Une infraction que n'ont pas manqué de relever de nombreux internautes. " Trois points et 135 euros d'amende ", note un utilisateur de Twitter. " Bonjour, les limitations de vitesse ne s'appliquent pas aux parlementaires ? ", ironise un autre.Opposant au 80km/heureCertains ont par ailleurs fait le lien entre les positions politiques de l'élu communiste et cet excès de vitesse. Car Jean-Paul Dufrègne est un farouche opposant au projet du gouvernement de limiter le réseau français de routes secondaires à 80 km/heure. Avec une trentaine d'autres élus du Massif Central, il avait d'ailleurs adressé une lettre ouverte à Emmanuel Macron sur le sujet, dénonçant une mesure " injuste et pénalisante ", et un frein au développement du Massif Central. \\ 
 \hline 
 \hline 
 \multirow{5}{*}[-3.5em]{\rotatebox[origin=c]{90}{\textsc{Abstract}}} & Gold & Jean-Paul Dufrègne, élu de l'Allier, a fait sensation sur les réseaux sociaux après être apparu dans un reportage de TF1. On le voyait rouler à plus de 120 km/heure sur une route secondaire. \\ 
 & mBART & L'élu communiste de l'Allier a été filmé le 4 avril au journal de 20 heures de TF1, comme le relève LCI. Son compteur affichait 124 km/heure au-dessus de la limite autorisée.\\ 
 & mBARThez &Son compteur affichait 124 km/heure au-dessus de la limite autorisée. \\ 
 & BARThez &Un élu communiste de l'Allier a été filmé le temps d'un reportage par les caméras de TF1, dans la soirée du 4 avril, alors qu'il roulait 90 km/heure au-dessus de la limite autorisée.\\ 
 & C2C & Un député communiste du Puy-de-Dôme n'a pas manqué de se faire remarquer en pleine séance de questions au gouvernement. Il était l'invité de Jean-Marc Remontini, l'élu apparenté du PCF, qui a fait l'éloge de\\ 
 \hline 
 \hline 
 \multirow{5}{*}[-2em]{\rotatebox[origin=c]{90}{\textsc{Title}}} & Gold &Un élu épinglé à 124 km/heure sur une route limitée à 90 \\ 
 & mBART &Un élu communiste flashé à 124 km/heure sur TF1\\ 
 & mBARThez &Un élu communiste flashé à 124 km/heure sur TF1\\ 
 & BARThez &Un élu communiste passe un sale quart d'heure à cause de son compteur de vitesse\\ 
 & C2C & Ce député du Doubs qui n'a plus le temps de répondre aux radars\\ 
 \hline 
 \end{tabular} 
 \caption{C2C stands for CamemBERT2CamemBERT. OrangeSum document 22077.} 
 
 \end{table*}

\begin{table*} 
 \centering 
 \begin{tabular}{|cr|L|} 
 \hline 
 & Document &\baselineskip=8pt {\small Mais où est donc passé Gérald Thomassin ? L'acteur français qui avait obtenu un César en 1991 est introuvable depuis le 28 août dernier, rapporte RTL. Le comédien âgé de 45 ans devait se rendre à un rendez-vous judiciaire dans une affaire de meurtre. Mais il ne s'y est jamais rendu. Et depuis, c'est toute sa famille qui s'inquiète. Interrogé par RTL, le frère de l'acteur, Jérôme Thomassin, a montré toute son inquiétude avant d'apporter des détails sur la journée du 28 août.Selon lui, Gérald Thomassin a bien "pris le train Rochefort-Lyon pour se rendre à la confrontation avec deux autres mis en examen". Parmi ces hommes, précise RTL, le principal suspect dans cette affaire de meurtre dans un bureau de poste. Les avocats du comédien qui appartiennent au cabinet d'Éric Dupond-Moretti ont signalé "une disparition inquiétante" au commissariat de Rochefort (Charente-Maritime) où l'acteur vivait. En tout état de cause, son frère était "très heureux de pouvoir se rendre à ce rendez-vous judiciaire." "L'affaire Burgod" L'affaire remonte à 2013, lorsque Gérald Thomassin est interpellé et mis en examen pour "vol avec arme et homicide sur une personne chargée d'une mission de service public". Une employée de La Poste, Catherine Burgod, enceinte, avait été tuée de 28 coups de couteau. Tenu responsable, l'acteur avait été incarcéré en 2013 avant d'être remis en liberté, mais placé sous contrôle judiciaire en octobre 2015. Sauf qu'il décide de briser son bracelet électronique et retourne en prison. Gérald Thomassin sort finalement en 2016, après trois ans de détention provisoire, la limite. L'affaire prend une autre tournure en 2017 et 2018 avec l'arrestation d'un suspect et la mise en examen d'un autre, mais la justice ne parvient toujours pas à trancher. La reconstitution du jeudi 29 août aurait dû permettre une confrontation entre les trois protagonistes, mais Gérald Thomassin ne s'est jamais présenté, au grand dam de l'avocate des parties civiles. Aujourd'hui, la famille de l'acteur demande de vraies recherches. Ses appels sonnent dans le vide. Ses proches craignent qu'il ait pu faire une mauvaise rencontre en cours de route. Le comédien avait reçu le César du Meilleur jeune espoir en 1991 pour son rôle dans "Le Petit Criminel" de Jacques Doillon, avant de connaitre une carrière ponctuée de plus ou moins grands rôles.} \\ 
 \hline 
 \hline 
 \multirow{5}{*}[-3.5em]{\rotatebox[origin=c]{90}{\textsc{Abstract}}} & Gold & Gérald Thomassin a disparu fin août selon ses proches qui s'inquiètent, révèle RTL. Il devait se rendre à un rendez-vous judiciaire pour éclairer une affaire d'homicide. \\ 
 & mBART & Gérald Thomassin a disparu fin août. Il devait se rendre à un rendez-vous judiciaire pour éclairer une affaire d'homicide, révèle RTL.\\ 
 & mBARThez &Gérald Thomassin a disparu fin août. Il devait se rendre, selon RTL, à un rendez-vous judiciaire pour éclairer une affaire de meurtre. \\ 
 & BARThez &Le comédien de 45 ans devait se rendre à un rendez-vous judiciaire dans une affaire de meurtre, mais il ne s'y est jamais rendu, rapporte RTL.\\ 
 & C2C & Gérald Thomassin a disparu fin août. Il se rendait à un rendez-vous judiciaire pour éclairer une affaire d'homicide.\\ 
 \hline 
 \hline 
 \multirow{5}{*}[-2em]{\rotatebox[origin=c]{90}{\textsc{Title}}} & Gold &Mystérieuse disparition d'un acteur césarisé soupçonné de meurtre \\ 
 & mBART &Disparition de l'acteur Gérald Thomassin : sa famille s'inquiète\\ 
 & mBARThez &L'acteur Gérald Thomassin porté disparu depuis le 28 août\\ 
 & BARThez &L'acteur Gérald Thomassin porté disparu depuis le 28 août\\ 
 & C2C & Disparition de l'acteur Gérald Thomassin : la famille n'est plus introuvable\\ 
 \hline 
 \end{tabular} 
 \caption{C2C stands for CamemBERT2CamemBERT. OrangeSum document 22168.} 
 
 \end{table*}

\begin{table*} 
 \centering 
 \begin{tabular}{|cr|L|} 
 \hline 
 & Document &\baselineskip=8pt {\small Dans un rapport adressé aux ministres de l'Intérieur, de la Justice, et à la secrétaire d'Etat à l'Egalité femmes-hommes Marlène Schiappa, les cinq députés chargés d'étudier la verbalisation du harcèlement de rue recommandent la mise en place d'"une contravention de 4e classe d'outrage sexiste et sexuel". L'infraction devra être constatée "en flagrance" par les agents de la toute récente "police de proximité du quotidien", précise leur texte, qui, selon les informations du Huffington Post, devrait être remis mercredi 28 février.Jusqu'à 1.500 euros d'amendesLe montant de l'amende forfaitaire serait de 90 euros pour un paiement immédiat, 200 euros pour un paiement sous 15 jours et 350 euros en peine majorée. En cas de circonstances aggravantes (si l'auteur est dépositaire de l'autorité publique, en cas de réunion, ou de bande organisée), une contravention de 5e classe (jusqu'à 1.500 euros) pourrait être délivrée par un tribunal de police.Pour Sophie Auconie (UDI, Agir et Indépendants), Laetitia Avia (LREM), Erwan Balanant (Modem), Elise Fajgeles (LREM) et Marietta Karamanli (Nouvelle gauche), le harcèlement subi dans l'espace public est un "fléau". Ils estiment nécessaire de "définir une nouvelle infraction visant à sanctionner, entre autres, les gestes déplacés, les sifflements, les regards insistants ou remarques obscènes, le fait de suivre volontairement à distance une personne créant ainsi une situation d'angoisse", soulignent-ils.68\% des Français favorables aux amendesLe rapport souhaite également que les auteurs participent à un stage de sensibilisation à l'égalité femmes-hommes, et que la police municipale et les agents des services de sécurité des transports soient habilités à constater cette infraction. D'après un sondage Opinionway réalisé pour Public Sénat, Les Echos et Radio Classique et publié le 5 février, une large majorité de Français est favorable à la mise en place d'une amende pénalisant le harcèlement de rue. À la question "êtes-vous favorable ou pas favorable à ce que le harcèlement de rue (sifflements, remarques...) soit passible d'une amende ?", 68 \% des personnes interrogées se disent favorables (40 \% "plutôt favorables" et 28 \% "tout à fait favorables"). 30 \% y sont opposés (23 \% "plutôt opposés" et 7 \% "tout à fait opposés") et 2 \% ne se prononcent pas.} \\ 
 \hline 
 \hline 
 \multirow{5}{*}[-3.5em]{\rotatebox[origin=c]{90}{\textsc{Abstract}}} & Gold & Des parlementaires préconisent de créer une infraction d'"outrage sexiste" sanctionnant d'une amende immédiate de 90 euros "tout propos, comportement ou pression à caractère sexiste ou sexuel" dans l'espace public. \\ 
 & mBART & Selon un rapport, dévoilé par le Huffington Post, le gouvernement envisage une amende forfaitaire de 90 euros pour lutter contre le harcèlement de rue. En cas de circonstances aggravantes, elle pourrait être délivrée par un tribunal de police.\\ 
 & mBARThez &Dans un rapport adressé aux ministres de l'Intérieur, de la Justice et à la secrétaire d'Etat à l'Egalité femmes-hommes, les députés chargés d'étudier la verbalisation du harcèlement de rue recommandent la mise en place d'une contravention de 4e classe. \\ 
 & BARThez &D'après un sondage Opinionway réalisé pour Public Sénat, Les Echos et Radio Classique, une large majorité de Français sont favorables à la mise en place d'une amende pénalisant le harcèlement de rue.\\ 
 & C2C & Selon un sondage Elabe pour Le Huffington Post, 54\% des Français sont opposés au projet de loi sur le harcèlement de rue. Une première en soi, alors que la question de l'emprise sexuelle se pose déjà : les contraventions seront en effet posées\\ 
 \hline 
 \hline 
 \multirow{5}{*}[-2em]{\rotatebox[origin=c]{90}{\textsc{Title}}} & Gold &Harcèlement de rue : bientôt une amende immédiate de 90 euros ? \\ 
 & mBART &Harcèlement de rue : vers une contravention de 4e classe ?\\ 
 & mBARThez &Harcèlement de rue : vers une contravention de 4e classe ?\\ 
 & BARThez &Harcèlement de rue : vers une contravention de 4e classe ?\\ 
 & C2C & Harcèlement de rue : un rapport préconise une amende de 5 à 5 euros\\ 
 \hline 
 \end{tabular} 
 \caption{C2C stands for CamemBERT2CamemBERT. OrangeSum document 22423.} 
 
 \end{table*}

\begin{table*} 
 \centering 
 \begin{tabular}{|cr|L|} 
 \hline 
 & Document &Le 18 octobre dernier, Jacline Mouraud se faisait connaître en publiant sur Facebook une vidéo dans laquelle elle poussait un "coup de gueule" contre le gouvernement. Aujourd'hui, la Bretonne a pris ses distances par rapport au mouvement, notamment face à d'autres figures plus radicales comme Éric Drouet.Jacline Mouraud réfléchit désormais à créer son propre parti, "la seule chose envisageable", comme elle l'explique au JDD.Nicolas Sarkozy, "le seul qui a des couilles"Cette figure des "gilets jaunes", accusée de faire le jeu de LREM estime que "le problème" d'Emmanuel Macron "c'est qu'il est jeune". "Il devrait y avoir un âge minimum pour être président : 50 ans", souligne Jacline Mouraud.Dans le JDD, elle raconte d'ailleurs avoir voté blanc lors de la dernière présidentielle. En 2007 et 2012, c'est Nicolas Sarkozy, "le seul qui a des couilles", que la figure des "gilets jaunes" avait soutenu. En attendant de se lancer, pas question pour elle en tous les cas d'être candidate aux européennes sur une liste de La République en marche. \\ 
 \hline 
 \hline 
 \multirow{5}{*}[-3.5em]{\rotatebox[origin=c]{90}{\textsc{Abstract}}} & Gold & L'une des figures du mouvement ne sera toutefois pas candidate aux prochaines élections européennes. \\ 
 & mBART & Jacline Mouraud, figure des "gilets jaunes", estime que le président d'Emmanuel Macron est trop jeune pour être président.\\ 
 & mBARThez &Dans un entretien au JDD, la figure des "gilets jaunes" Jacline Mouraud révèle qu'elle réfléchit à créer son propre parti. \\ 
 & BARThez &Dans les colonnes du JDD, la figure des "gilets jaunes" explique qu'elle envisage de se présenter aux européennes sur une liste La République en marche.\\ 
 & C2C & Retirée de la vie politique depuis plusieurs mois, Bretone Mouraud envisage de se lancer en politique. Et elle réfléchit à quelque chose de plus, rapporte le JDD.\\ 
 \hline 
 \hline 
 \multirow{5}{*}[-2em]{\rotatebox[origin=c]{90}{\textsc{Title}}} & Gold &"Gilets jaunes" : Jacline Mouraud réfléchit à créer son parti \\ 
 & mBART &"Gilets jaunes" : Jacline Mouraud lance son propre parti\\ 
 & mBARThez &"Gilets jaunes" : Jacline Mouraud prend ses distances\\ 
 & BARThez &La figure des "gilets jaunes" Jacline Mouraud va créer son propre parti\\ 
 & C2C & "Gilets jaunes" : Jacline Mouraud réfléchit à sa propre candidature\\ 
 \hline 
 \end{tabular} 
 \caption{C2C stands for CamemBERT2CamemBERT. OrangeSum document 19233.} 
 
 \end{table*}

\begin{table*} 
 \centering 
 \begin{tabular}{|cr|L|} 
 \hline 
 & Document &\baselineskip=8pt {\small Invité du "Grand rendez-vous Europe 1/CNews/Les Échos dimanche 8 avril, Jean-Luc Mélenchon a appelé "à faire baisser la température dans ce pays". En cause : les menaces de mort dont il ferait l'objet, ainsi que d'autres élus LFI. Le député des Bouches-du-Rhône a confirmé avoir récemment demandé que le ministre de l'Intérieur Gérard Collomb soit entendu dans l'enquête sur un projet d'attentat d'ultra-droite où il a été cité comme cible potentielle. "Je me suis porté partie civile dans cette affaire. J'ai appris en octobre dernier qu'un groupe de gens avait l'intention de me tuer, ainsi que (le secrétaire d'État) M. Castaner". Or pendant la campagne législative de juin 2017, "j'ai demandé à être protégé" car "j'avais reçu à Marseille des menaces de mort. On me l'a refusé, et puis après je découvre que le 28 mai, ils ont arrêté ce personnage (...) Quatre mois plus tard ils en arrêtent neuf autres qui étaient toujours en action pendant ces quatre mois". "LA RECRUDESCENCE D'UN EXTRÉMISME D'EXTRÊME DROITE EXTRÊMEMENT VIOLENT"Un ancien militant du groupuscule royaliste Action française en Provence, Alexandre Nisin, a été mis en examen début juillet pour association de malfaiteurs terroriste criminelle. Huit autres suspects ont été mis en examen, soupçonnés d'appartenir à son réseau. "Ni moi, ni Castaner n'avons été prévenus de rien", a déploré l'ancien candidat à la présidentielle. "Sur 17 que nous sommes au groupe La France insoumise (à l'Assemblée, ndlr), il y en a cinq qui font l'objet de menaces de mort"(/BOLD], a-t-il par ailleurs révélé. Jean-Luc Mélenchon a dénoncé "la recrudescence d'un extrémisme d'extrême droite extrêmement violent, dans toutes sortes de villes, qui va jusqu'à des tentatives d'assassinat". "L'extrême droite doit être prise au sérieux comme danger de violence et de meurtre. C'est eux qui attaquent à Montpellier un amphi d'étudiants, c'est eux qui attaquent à Tolbiac, c'est eux qui me menacent de mort. C'est eux qui font des contrôles d'identité dans la rue dans au moins deux villes. Ça suffit. Maintenant le ministre de l'Intérieur doit prendre au sérieux la menace que représentent les groupuscules radicalisés de l'extrême droite", a-t-il poursuivi."Il y a des groupes d'extrême droite qui prolifèrent dans le pays. Qui souvent ont commencé leurs premiers pas avec le Front national et qui maintenant vont au bout de cette logique", a-t-il ajouté.} \\ 
 \hline 
 \hline 
 \multirow{5}{*}[-3.5em]{\rotatebox[origin=c]{90}{\textsc{Abstract}}} & Gold & Le leader de La France insoumise (LFI) dénonce la "recrudescence" d'une "extrême droite extrêmement violent(e)" en France, qui doit être "prise au sérieux" par le gouvernement. \\ 
 & mBART & S'il dénonce la recrudescence d'un extrémisme d'extrême droite "extrêmement violent" dans certaines villes, le chef de file de La France insoumise (LFI), Jean-Luc Mélenchon, s'est attaqué au ministre de l'Intérieur, Gérard Collomb.\\ 
 & mBARThez &Au micro d'Europe 1 dimanche 8 avril, Jean-Luc Mélenchon a réagi aux menaces de mort dont il fait l'objet et dénoncé "la recrudescence d'un extrémisme d'extrême droite extrêmement violent". \\ 
 & BARThez &- Le chef de file de La France insoumise et ancien candidat à la présidentielle est vent debout contre le projet d'attentat déjoué à Marseille. Il estime que le ministre de l'Intérieur, Gérard Collomb, est menacé de mort par un groupe d'extrême droite.\\ 
 & C2C & Selon le leader de La France insoumise (LFI), le député des Bouches-du-Rhône, Jean-Luc Mélenchon, "rappelle à tous ceux suspectés d'avoir menacé d'assassiner le ministre de l'Intérieur, ce que conteste le parti et\\ 
 \hline 
 \hline 
 \multirow{5}{*}[-2em]{\rotatebox[origin=c]{90}{\textsc{Title}}} & Gold &VIDÉO. Cinq députés de La France insoumise font l'objet de menaces de mort, selon Jean-Luc Mélenchon \\ 
 & mBART &Menaces de mort : Jean-Luc Mélenchon s'en prend à Castaner\\ 
 & mBARThez &Jean-Luc Mélenchon dénonce les "menaces de mort" de Gérard Collomb\\ 
 & BARThez &VIDÉO. Jean-Luc Mélenchon dénonce les menaces de mort dont Gérard Collomb est victime\\ 
 & C2C & Projet VIDÉO. Menace de mort à Marseille : Jean-Luc Mélenchon menace de démissionner\\ 
 \hline 
 \end{tabular} 
 \caption{C2C stands for CamemBERT2CamemBERT. OrangeSum document 22060.} 
 
 \end{table*}

\end{document}